\documentclass[letterpaper, 10 pt, conference]{ieeeconf}  % Comment this line out
\IEEEoverridecommandlockouts                              % This command is only needed if you want to
\usepackage[utf8]{inputenc}
\usepackage{amsmath,amssymb,stmaryrd,mathtools}

\usepackage{amsfonts}
\usepackage{breqn}

\allowdisplaybreaks

\usepackage{algorithm}
\usepackage[noend]{algpseudocode}
\usepackage{acronym}
\usepackage{color,xcolor}
\usepackage{verbatim}
\usepackage{booktabs}
\usepackage{siunitx}
\usepackage{graphics}
\usepackage{graphicx,caption}
\usepackage{suffix}
\usepackage{xstring}
\usepackage{xparse}
\usepackage{expl3}
\usepackage{mathrsfs}
\usepackage{tabularx}
\usepackage{makecell}
\usepackage{array}
\usepackage{hyperref}
\usepackage{cleveref}
\usepackage{multirow}
\usepackage{diagbox}
\usepackage{rotating}
\usepackage{subcaption}
\usepackage{wrapfig}

\usepackage{tikz}
\usetikzlibrary{arrows,backgrounds,calc}
\usepackage{relsize}
\usepackage{float}
\usepackage{kantlipsum} 
\usepackage{lipsum}
\usepackage{stfloats}
\usepackage{siunitx}

\usepackage[noadjust]{cite}
\usepackage{todonotes}
\usepackage{soul}
\definecolor{smoothgreen}{rgb}{0.7,1,0.7}
\sethlcolor{smoothgreen}

\RequirePackage{luatex85}
\usepackage{pgfplots}
\pgfplotsset{compat=newest}
\pgfplotsset{every axis legend/.append style={%
		cells={anchor=west}}
}
\usepgfplotslibrary{polar}
\usetikzlibrary{arrows}
\tikzset{>=stealth'}

\definecolor{C1}{rgb}{0.0, 0.447, 0.741}
\definecolor{C1_light}{rgb}{0.0, 0.6032388663967612, 1.0}
\definecolor{C2}{rgb}{0.85, 0.325, 0.098}
\definecolor{C3}{rgb}{0.929, 0.694, 0.125}
\definecolor{C4}{rgb}{0.494, 0.184, 0.556}
\definecolor{C5}{rgb}{0.466, 0.674, 0.188}
\definecolor{C6}{rgb}{0.301, 0.745, 0.933}
\definecolor{C7}{rgb}{0.635, 0.078, 0.184}

\usepgfplotslibrary{groupplots}

\usetikzlibrary{shapes.geometric, arrows}

\tikzstyle{startstop} = [rectangle, rounded corners, minimum width=2cm, minimum height=1cm,text centered, draw=black, fill=none]
\tikzstyle{arrow} = [thick,->,>=stealth]

\title{%\LARGE \bf 
User-Friendly Safety Monitoring System for Manufacturing Cobots}

\author{ Ye-Ji Mun$^*$, Zhe Huang$^*$, Haonan Chen, Yilong Niu, Haoyuan You,\\ D. Livingston McPherson, and Katherine Driggs-Campbell
\thanks{
$^*$ denotes equal contribution. 
}%
\thanks{Y. Mun, Z. Huang, H. Chen, Y. Niu, H. You, D. McPherson, and K. Driggs-Campbell are with the Department of  Electrical and Computer Engineering at the University of Illinois at Urbana-Champaign. emails: \{yejimun2, zheh4, haonan2, yilongn2, hy19, dlivm, krdc\}@illinois.edu}
\thanks{We thank Foxconn Interconnect Technology for supporting the authors with their research. This work was supported in part by ZJU-UIUC Joint Research Center Project No. DREMES 202003, funded by Zhejiang University.}
}

\begin{document}
\bstctlcite{IEEEexample:BSTcontrol}
\maketitle
\thispagestyle{empty}
\pagestyle{empty}

%%%%%%%%%%%%%%%%%%%%%%%%%%%%%%%%%%%%%%%%%%%%%%%%%%%%%%%%%%%%%%%%%%%%%%%%%%%%%%%%
\begin{abstract}
Collaborative robots are being increasingly utilized in industrial production lines due to their efficiency and accuracy. However, the close proximity between humans and robots can pose safety risks due to the robot's high-speed movements and powerful forces. To address this, we developed a vision-based safety monitoring system that creates a 3D reconstruction of the collaborative scene. Our system records the human-robot interaction data in real-time and reproduce their virtual replicas in a simulator for offline analysis. The objective is to provide workers with a user-friendly visualization tool for reviewing performance and diagnosing failures, thereby enhancing safety in manufacturing settings. 
\end{abstract}

% %%%%%%%%%%%%%%%%%%%%%%%%%%%%%%%%%%%%%%%%%%%%%%%%%%%%%%%%%%%%%%%%%%%%%%%%%%%%%%%%
\section{Introduction}
\label{sec:introductions}

The automation of industrial robots in the new paradigm of Industry $4.0$ has increased the potential of robots operating near human workers. In such settings, a combination of the human's adaptability in dynamic environments and the robot's high accuracy in repetitive tasks can increase productivity in manufacturing assembly lines. However, industrial robots can be dangerous for human workers to work with in close proximities due to high-speed movements and massive forces. Wherever human and industrial robots share a common workplace, accidents are likely and always unpredictable. In such settings, a robust safety system to monitor potential accidents and analyze the robot's safety system is crucial. 

A stream of works has been introduced to secure safe human-robot collaboration in a shared workspace. One of the most common vision-based methods for ensuring human safety is regulating robot motion based on its distance from human workers~\cite{halme2018review}. In these approaches, the robot reduces its velocity or force when the humans intrude into the safety zone, where the human and the robots are at a close distance and at a safety critical status~\cite{kang2019safety}. Some works limit the force or power of the robot to reduce the severity of human injury when collisions happen~\cite{standard2016iso}. The recent advanced methods reconstruct and track the working environment to monitor safety-related sensor data in its digital twin world. Such digital twin systems provide real-time analysis and simulation for safety monitoring and production management~\cite{baidya2022digital}. However, accidents can still occur, and to prevent such incidents, monitoring the safety issue by reproducing the accident in the virtual environment can help inspectors better understand the collaborative workflow and diagnose any safety anomalies in the robot systems. 

In this work, we designed a vision-based safety monitoring system that allows real-time and offline monitoring for safety analysis. Our safety monitoring system tracks safety critical period when the human's body parts intrude the safety zone. Our system records the real-time human-robot interaction data and visualizes the solution process execution to critique the workflow offline in a 3D scene reproduction. We employ a graphical user interface (GUI) to integrate the process and make it easier for the engineers to deploy our system. Note that our system is different from the digital twin (as per the taxonomy in Baidya et al.~\cite{baidya2022digital}) since our digital environment visualizes the scene and does not influence the physical model itself. However, by adopting this approach, it becomes simpler to identify safety concerns in a robot's planning and workflow.

\section{Methodology}
\label{sec:methodology}

\begin{figure}[t]
\centering
\includegraphics[width=0.45\textwidth]{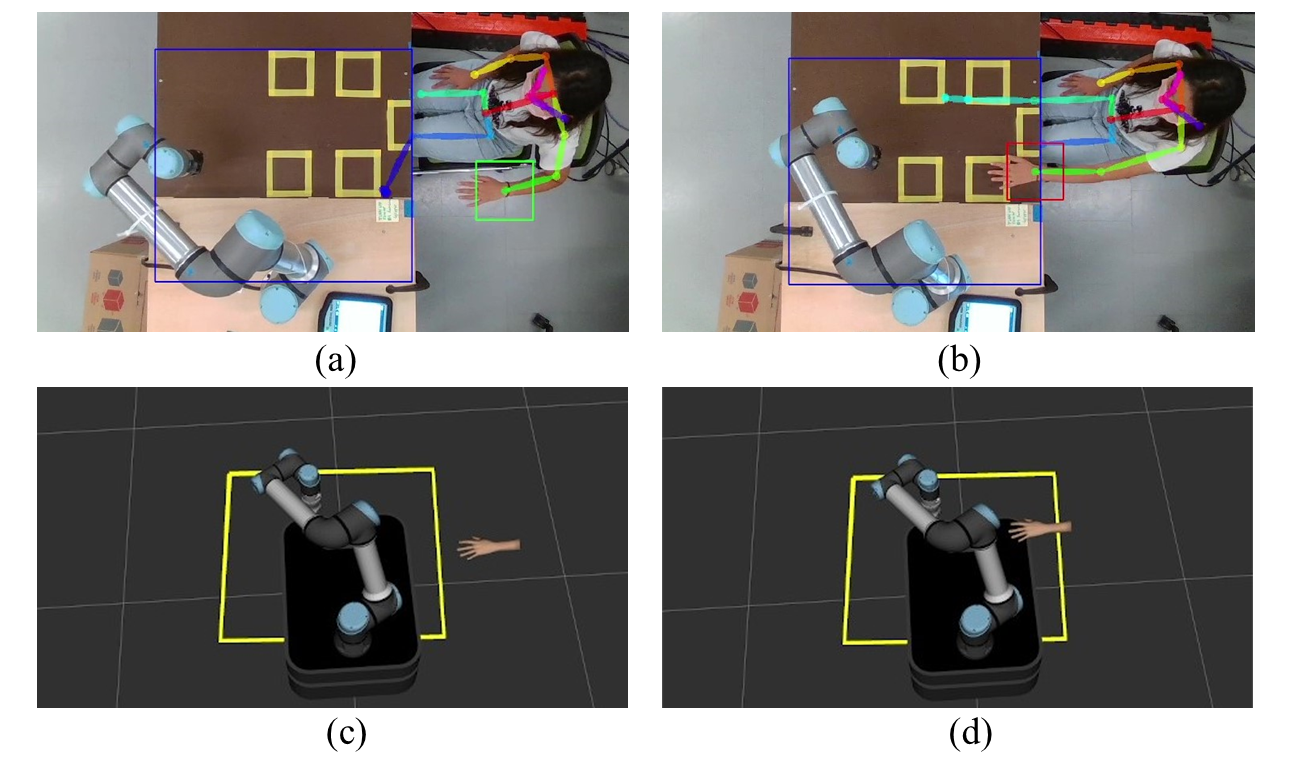}
\vspace{-0.1cm}
\caption{Photographs of our experimental setup (a, b) with corresponding scene visualizations (c, d). As the human reaches the safety critical zone (b, d), a safety flag is turned on to denote the safety critical period; the safety flag is off when the human is out of the safety critical zone (a, c).} 
\label{fig:intro}
\vspace{-0.6cm}
\end{figure}

Our vision-based safety monitoring system achieves the general industrial safety requirements: real-time monitoring, recording, and reproduction of the collaborative manufacturing process. Our system collects real-world data from sensors and reconstructs its digital counterpart in $3$D simulator, which allows workers to review performance and diagnose any failings through easy-to-deploy visualizations. To enable this, we build a simple launcher that integrates disparate software pieces into a unified user interface in NodeRed.

%% Real-time Visualization 
\paragraph{Vision-based Safety Monitoring}

Our safety monitoring system tracks the safety critical periods during human-robot collaboration for safety analysis. We first predefine the shared workspace, a safety critical zone, and monitor whether the human body part is within this safety zone. When the human worker intrudes into the space, a safety flag is turned on to denote the safety-critical period. Our system visualizes and records the safety data and the robot and human motion in real-time. Our system allows factory operators to monitor the robot alongside safety-critical human positions to prevent time-critical accidents, and provides more reliable scene replay without suffering from latency. We replicate the dynamic real-world environment in RViz based on the sensor data collected by the robot and a top-down view camera. We use OpenPose
% ~\cite{cao2017realtime} 
to estimate human pose from the camera image. Figure~\ref{fig:intro} shows the comparison between the real world (a, b) and the scene visualizer (c, d). The blue box in the camera view represents the shared workspace, which corresponds to the yellow rectangle in the Rviz. We track whether the human's left wrist is within the shared workspace (denoted by a red rectangle in (b)) or not (denoted by a green rectangle in (a)), which determines the safety critical period. Our digital twin safety system achieves both real-time and offline safety monitoring.

The scene replay module stores recorded data in YAML format to enable real-time data retrieval, control of playback speed, and easy inspection of sensor data at any timestamp by investigators. In order to capture both detailed and high-level information from the robot task execution process, we implement three modules in the replay system: the joint replay module, the hand replay module, and the scene replay module. The joint replay module publishes the position of the robot joints. The hand replay module uses back projection to project the pixel coordinates of the detected hand to its 3D coordinates. We use camera extrinsic calibration to obtain the extrinsic matrices (with respect to the base link of the robot arm) for the back projection. The scene replay module is the integration of the joint replay module and the hand replay module and replays the recorded data in the YAML file to RViz using functions defined in the two modules. The visualizing frame rate is $20$ Hz.

\paragraph{Graphical User Interface}

\begin{figure}[t!]
    \centering
    \includegraphics[width = 0.18\textwidth]{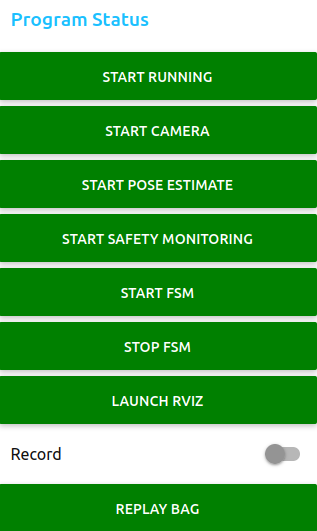}
    \caption{A GUI designed for our user-friendly safety monitoring system.}
    \label{fig:user_interface}
    \vspace{-0.6cm}
\end{figure}

Integrating the safety monitoring in the digital twin with the extended tasks requires a unified graphical user interface (GUI) to launch our disparate functions such as robot execution, safety monitoring, recording, and scene replay. Figure~\ref{fig:user_interface} shows the design of our GUI. Our user interface initiates the robot (`START RUNNING`), executes planned motion for the task (`START FSM`), and stops the motion (`STOP FSM`) by pressing user-friendly buttons on our interface. The vision-based safety monitoring system can be launched by starting camera (`START CAMERA'), running OpenPose (`START POSE ESTIMATE`), and tracking the safety-critical period (`START SAFETY MONITORING`). The factory operator can start/stop recording the sensor data anytime using a slider (`Record`). For visualization in the virtual world, RViz needs to be launched (`LAUNCH RVIZ`), and the scene can be reproduced (`REPLAY BAG`) based on the last recording. These functionalities are compactly integrated into our GUI, and our GUI is designed to easily replace the robot planning algorithms, and other functionalities. For example, to change the robot task, the planning script connected to the robot planning button can be swapped. As a consequence, our GUI program allows users to operate the safety monitoring system independent of the robot task execution programs and has the potential to be expanded to more complex tasks in the future.

% \section{Conclusion and Future Works}

% We present a vision-based safety monitoring system that enables real-time and offline safety analysis. During human-robot collaboration, our system monitors safety-critical time periods and records interaction data, which can be then reproduced in a 3D digital environment for workflow analysis. To facilitate user-friendly deployment, we utilize a graphical user interface (GUI) to integrate the system's functions. Our framework is designed to easily adapt different algorithms and functions. 
\section{Future Works}

One potential avenue for extending our safety monitoring system is to facilitate the transfer of virtually acquired knowledge to optimize real-world operational skills, similar to a digital twin. Rather than relying on manual analysis and optimization, the augmented data obtained from the simulator can help find better control strategies. For instance, possible trajectories of the robot can be simulated to check the potential collision, and the best safe route can be selected based on predicted human trajectories. One of the challenges to address in this setup is human modeling in the simulation. We have implemented a real-time human motion prediction pipeline that extracts human poses from raw image frames and converted to 3D coordinates using depth information. The video is available at \url{https://youtu.be/RVSWExpYO6o}. Integrating our safety monitoring system and human motion prediction pipeline to build a safety digital twin system can alleviate the limitations of our current system and enhance the safety of all participants involved. 

\label{sec:conclusion}

% \section*{ACKNOWLEDGMENT}
% This work utilizes resources supported by the National Science Foundation’s Major Research Instrumentation program, grant $\#1725729$, as well as the University of Illinois at Urbana-Champaign~\cite{10.1145/3311790.3396649}.

%%%%%%%%%%%%%%%%%%%%%%%%%%%%%%%%%%%%%%%%%%%%%%%%%%%%%%%%%%%%%%%%%%%%%%%%%%%%%%%%

% \addtolength{\textheight}{-12cm}   % This command serves to balance the column lengths
                                  % on the last page of the document manually. It shortens
                                  % the textheight of the last page by a suitable amount.
                                  % This command does not take effect until the next page
                                  % so it should come on the page before the last. Make
                                  % sure that you do not shorten the textheight too much.

%%%%%%%%%%%%%%%%%%%%%%%%%%%%%%%%%%%%%%%%%%%%%%%%%%%%%%%%%%%%%%%%%%%%%%%%%%%%%%%%

% \newpage
% \clearpage
\bibliographystyle{IEEEtran}
\bibliography{BibFile}
\clearpage
\end{document}